\documentclass[letterpaper]{article} 
\usepackage{aaai25}  
\usepackage{times}  
\usepackage{helvet}  
\usepackage{courier}  
\usepackage[hyphens]{url}  
\usepackage{graphicx} 
\usepackage{natbib}  
\usepackage{caption} 
\usepackage{amsmath}
\usepackage{amssymb}
\usepackage{algorithm}
\usepackage{algorithmic}

\usepackage[table]{xcolor}
\definecolor{lightgreen}{rgb}{0.9,1,0.9}

\usepackage{newfloat}
\usepackage{listings}
\DeclareCaptionStyle{ruled}{labelfont=normalfont,labelsep=colon,strut=off} 
\floatstyle{ruled}
\newfloat{listing}{tb}{lst}{}
\floatname{listing}{Listing}
\title{Evaluating Binary Decision Biases in Large Language Models: Implications for Fair Agent-Based Financial Simulations}
\author{
    Alicia Vidler, Toby Walsh 
}    
    
\affiliations{

    UNSW Sydney, Australia\\
    a.vidler@unsw.edu.au, t.walsh@unsw.edu.au 
}

\begin{document}
\maketitle

\begin{abstract}

Large Language Models (LLMs) are increasingly being used to simulate human-like decision making in agent-based financial market models (ABMs). As models become more powerful and accessible, researchers can now incorporate individual LLM decisions into ABM environments. However, integration may introduce inherent biases that need careful evaluation. In this paper we test three state-of-the-art GPT models for bias using two model sampling approaches: one-shot and few-shot API queries.  We observe significant variations in distributions of outputs between specific models, and model sub versions, with GPT-4o-Mini-2024-07-18 showing notably better performance (32-43\% yes responses) compared to GPT-4-0125-preview's extreme bias (98-99\% yes responses).  We show that sampling methods and model sub-versions significantly impact results: repeated independent API calls produce different distributions compared to batch sampling within a single call. While no current GPT model can simultaneously achieve a uniform distribution and Markovian properties in one-shot testing, few-shot sampling can approach uniform distributions under certain conditions.  We explore the Temperature parameter, providing a definition and comparative results. We further compare our results to true random binary series and test specifically for the common human bias of Negative Recency - finding LLMs have a mixed ability to 'beat' humans in this one regard. These findings emphasise the critical importance of careful LLM integration into ABMs for financial markets and more broadly.



\end{abstract}


\section{Introduction}

Large Language Models (LLMs) are founded on human text, and inherit patterns from their training data.  They demonstrate capabilities in understanding and generating human-like text across various domains \cite{chang2023survey} and replicating human like behaviours as agents \cite{park2023generative}. Since humans demonstrate well-documented biases in their perception of randomness \cite{Warren2018}, we test to see if LLMs produce or replicate these biases when asked to make probabilistic decisions. LLMs have been found to deviate from intended trading instructions \cite{vidler2024tradertalkllmbehaviouralabm}, leading to systematic decision making biases despite demonstrating LLMs' potential for multi-agent systems in finance.  This becomes particularly relevant when using LLMs to make binary choices (e.g. "Randomly choose yes or no") such as in financial markets trading applications.  If financial models or markets seek to replace random generating functions with LLMs - what would happen?

In financial markets, randomness is a widely used mechanism for safety and fairness: markets such as the London Stock Exchange open and close at a random time within a pre-set window\footnote{https://www.londonstockexchange.com/discover/news-and-insights/what-auction}, and many banks are regulated under rules such as BIS Basel IV, focused on estimation of risk and capital adequacy through use of underlying probability distributions\footnote{https://www.bis.org/bcbs/index.htm?m=88}. Throughout financial pricing and risk models, extensive reliance on distributional assumptions and sampling is common place \cite{hull2022options}.  Any integration of LLM's into ABMs, or more broadly, financial market trading or risk systems, will necessarily incorporate any LLM bias. 

In this paper we pose a deeper question about the integration of LLMs into financial models; give that random numbers form an integral part of financial markets modelling, and decision making is crucial, can state-of-the-art LLMs reliably reproduce a simple uniform probability distribution, producing fair binary choices in response to independent agent-like API requests?  Can an LLM "toss a fair coin" when we ask it to within a model?

\subsection{Our Contribution}
In this paper we explore the significant variations in coin toss experimental results achieved with three widely-used LLMs: GPT-3.5 Turbo-0125, GPT-4-0125-preview, and GPT-4o-mini-2024-07-18.  To mirror natural language and avoid potential usage bias with words such as "heads" or "tails", we define a coin toss as asking for a "yes" or "no" binary output.  We focus our results on "yes" and "no" decisions to increase applicability and broaden possible usage.  We build on recent work on randomness and human bias in LLMs by \cite{vankoevering2024randomrandomevaluatingrandomness}, which found a mixed picture of results. We extend their work to explore specific released versions of common LLM's and differences between sub versions.  We extend \cite{Renda2023} with new terminology to better reflect the way in which a typical ABM might approach querying an LLM: one-shot and few-shot sampling, likewise extending \cite{vankoevering2024randomrandomevaluatingrandomness} further also.  These query methods reflect sampling of a model state either independently (One-Shot) or simultaneously (Few-Shot).  Our analysis reveals significant deviations from uniform probability distributions in LLM binary decision making (see Tables \ref{tab:OneShot_chi_square_tests}, \ref{tab:FewShot_Avg_chi_square_tests_part2}), and response sequences show strong decision dependence (Tables \ref{tab:OneShot_markov_tests}, \ref{tab:fEWsHOT_Avg_markov_tests_batch}).   For ABM integration, one-shot queries using GPT-4o-Mini-2024-07-18 showed minimal bias despite non-Markovian output.  Few-shot sampling (100 responses per API call) approached uniform distribution through sample averaging, though non-Markovian.  We explore few-shot samples as distinct batches, which again show non uniform distributions but over 58\% of batches were Markovian in their response.  We also examine the Temperature parameter's impact on distributional properties across six settings in the vein of \cite{Peeperkorn2024} and \cite{vankoevering2024randomrandomevaluatingrandomness} but extend their work by testings specific versions with specific sampling regimes. Even with optimised temperature settings, no model achieves both uniform distribution and Markovian outputs (Figure \ref{fig:Temp_4oNon linearresults}).  We observe differences between specific GPT model subtypes, with very large variations in outputs.

To gain a context on natural language bias, we perform a sampling of the Common crawl \cite{commoncrawl} as recent research by \cite{tessema2024unifiedcrawlaggregatedcommoncrawl} has found that it can provide a valuable data source for fine tuning LLMs. Using the cut-off date of GPT-4o-mini-2024-07-18 model training (corresponding to the most SOTA model), we find naturally occurring Yes/No responses deviate substantially from uniform distribution.  Recent work by \cite{Harrison} points to GPT-3.5-turbo-0125 being capable in equivalent few shot tests to avoid "repetitive and sequential patterns compared to humans" paving the way for the possibility that LLM's might be closer to randomness than expected.  We extend and expand this work and contribute to the area by looking at binary series, across three SOTA models and two methods of API generation. 
 We explore these features to extend work such as \cite{vidler2024modellingopaquebilateralmarket} and \cite{vidler2024tradertalkllmbehaviouralabm}, where including LLM's in financial market simulation holds great promise of increase model fidelity and nuance.


\section{Recent and Relevant Literature on LLMs}

\subsection{SOTA Models}
In November 2024, OpenAi, the creators of ChatGPT, announced they are working on an autonomous agent based version of their seminal LLMs, reportedly called "Operator" \cite{OpenAIAgents2024}.  Agentic AI tools are at the forefront of LLM design \cite{NgA}. Pre-trained LLMs, such as GPT-3 \cite{bubeck2023sparks} and GPT 4 \cite{Luo2023}, have demonstrated proficiency in tasks such as dialogue generation and natural language interactions, however a fundamentally different questions is asked of an LLM in an agentic AI tool and ABM. Unlike text and analysis needs, ABMs rely on \textit{agency}, that is to say, agents are imbued, and defined by, the requirement to make a decision in some frame or context \cite{Wooldridge_2009}.  Many such implementations ultimately require a binary choice from the agent: to more or not to move - "yes" or "no".  While LLMs offer promising capabilities as decision making agents, they face both engineering implementation challenges \cite{Chopra2024} and fundamental issues with decision making processes that can significantly impact overall behaviour \cite{vidler2024tradertalkllmbehaviouralabm}.

\subsection{Generative Agent based models (GABMs)}
Current thinking suggests that integrating LLM's into multi-agent frameworks has great potential for future AI systems \cite{NgA}.  Initial work by \cite{park2023generative} found that generative agents could be realistic simulacra of humans. Early attempts to introduce an LLM as an agent \cite{Chopra2024} found granular agent decisions were unfeasible to implement from a technological standpoint .  More recent work on these GABMs in financial trading decisions, using GPT-4o-Mini-2024-07-18, showed realistic representations of order-to-trade ratios found in US equity markets \cite{vidler2024tradertalkllmbehaviouralabm}.

\subsection{Financial market use and Markov property}

When using an LLM as a decision maker within an ABM or any other model, it is foreseeable that there may be challenges in relation to the sampling of the LLM's internal distribution of tokens, in effect producing an LLM model specific probability distribution of responses.  We focus also on the question of binary outcomes being Markovian, or memory-less. The Markov property is a key financial market concept introduced into modelling of markets by \cite{Fama1965RandomWI}.  In ABM terms this translates into a need to have each agents decision be based only on the decision process at the time step in question. Non-Markovian properties of ABMs for financial decisions would point to information leakage, collusion or agent group dynamics by default, defeating the goal of independent agents as a starting point for modelling.  

The potential to utilise LLMs to enrich other model methods with human characteristics (such as human biases) is beginning to be explored in text based arenas such as chat bots \cite{Gu2024} and survey participation \cite{Tjuatja2024}, and human "sims" \cite{park2023generative} to name just a few.  With the rapid advancement of both foundational LLM models, and hardware, it is now feasible to incorporate LLMs into ABMs directly \cite{vidler2024tradertalkllmbehaviouralabm} without the need for archetypes, or reduced agent fidelity, something that as little as 6 month ago was not considered computationally feasible \cite{Chopra2024}.

\subsection{Human Bias in Random sequence generation}
Research into human capacity for generating random sequences spans decades in psychological literature \cite{GinsburgN;Karpiuk1994}, \cite{Towse1998}. Traditional findings identified three primary biases in human-generated random sequences: cycling (avoiding recently used numbers), seriation (stereotypical patterns like ascending or descending sequences), and repetition avoidance \cite{Angelike2024}.
Recent studies have revealed a more nuanced understanding of these limitations \cite{Warren2018}, \cite{Wong2021}. Researchers have identified key metrics for distinguishing human-generated sequences from truly random processes, notably "algorithmic complexity" and "repetition median" which are found to be especially effective for sequences under 20 repetitions \cite{Angelike2024}. As specific bias, \textit{Negative Recency} has been identified,where humans rely on recent responses in sequential decision-making (such as coin tosses) to influence subsequent choices (\cite{BaenaMirabete2023AMT}, \cite{vankoevering2024randomrandomevaluatingrandomness}) illustrating a "negative recency" bias.  Human psychology research has found that this effect varies with age \cite{BaenaMirabete2023AMT} due to differences in working memory capacity, necessitating age-based criteria in comparative analyses. Additionally, \cite{Gauvrit2017} demonstrated that age is the primary factor affecting random sequence complexity, while \cite{Biesaga2021} found that fatigue negatively impacts performance. Our research extends these insights focusing on the negative recency bias in random sequence generation \cite{Towse1998}, \cite{Angelike2024} in specific LLM versions and independent sampling methods.

\subsection{Computational Randomness}
Research on LLM randomness has revealed limitations in number generation \cite{Renda2023, Harrison, Liu2024}, probabilistic text \cite{Tjuatja2024, Renda2023, Peeperkorn2024}, and behavioral simulations \cite{Gu2024}. Studies show LLMs struggle with sampling from specific distributions \cite{Imani2023, Renda2023}, though findings conflict: \cite{Tjuatja2024} found greater bias than humans, while \cite{Harrison} demonstrated superior randomness in gpt-3.5-turbo-0125. Human comparisons often rely on \cite{Figurska2008}'s limited 37-person study, leaving room for more comprehensive analysis.

Our work addresses the understudied impact of temperature settings on LLM randomness by systematically analyzing its effects across three specific GPT model versions using a binary decision test.

\subsection{Temperature}
Temperature, while absent from \cite{vaswani2023attentionneed}, is a crucial parameter in modern LLM APIs that controls output randomness. Defaulting to 1 (max 2) across tested models, its name derives from the SoftMax function's connection to Boltzmann's thermodynamic work \cite{Bridle}, making it relevant for analysing decision-making behavior. We provide a formal definition here: \\

\textbf{Definition: SoftMax function} \( \sigma \colon \mathbb{R{^P}} \to (0,1){^P} \) , where $P \geq 1$, takes a vector $\mathbf{z} = (z_1, \ldots, z_K) \in \mathbb{R}^P$ and computes each component of vector $\sigma(\mathbf{z}) \in (0,1)^P$ with

        \begin{equation}
        \sigma(\mathbf{z})_i = \frac{e^{z_i}}{\sum_{j=1}^P e^{z_j}}.
        \label{Softmax}
        \end{equation}

With the addition of a scaler \(\beta\) to the exponential terms in Equation (\ref{Softmax}) produces; 
   \begin{equation}
        \sigma(\mathbf{z})_i = \frac{e^{\beta z_i}}{\sum_{j=1}^P e^{\beta z_j}}.
        \label{TempSoftmax}
    \end{equation}

\textbf{Definition: Temperature} Let \textbf{T} to be a value proportional to the inverse of \(\beta\) such that: \( \mathbb{T}= \frac{1}{\beta} \). 

Temperature's impact has been studied across LLM applications including creative writing \cite{Peeperkorn2024}, question answering \cite{renze2024effectsamplingtemperatureproblem}, coding \cite{Zhu_2023}, and structured reasoning \cite{Ouyang_2023, Hickman_2024}. While \cite{Wang_2024} found task-dependent optimal temperatures, methodological approaches vary: \cite{Harrison} used only model defaults, whereas \cite{vankoevering2024randomrandomevaluatingrandomness} tested multiple settings but omitted model specifics. Our work extends this field by examining Temperature's specific impact on a binary decision distribution, independently sampled for three specific GPT sub-versions.

\section{Testing methodology}
We test LLM binary decision making, across three OpenAI models and refer to the following \textbf{models} (\footnote{details taken from API documentation https://platform.openai.com/docs/modelsgpt-4o-mini}):

\begin{itemize}
    \item \textbf{M1} (gpt-4o-mini-2024-07-18): Advanced small model optimised for cost-efficiency
    \item \textbf{M2} (gpt-4-0125-preview): Full GPT-4 Turbo with enhanced instruction following
    \item \textbf{M3} (gpt-3.5-turbo-0125): Latest GPT-3.5 iteration with improved format adherence
\end{itemize}

We extend Few-Shot Learning Terminology to LLM Querying. We borrow from \cite{vinyals2017matchingnetworksshotlearning} and present methods similar in the vein of \cite{Renda2023} and define specific query methods:  

\textbf{Definition: One-Shot Querying}:
\begin{description}
    \item{Independent API calls with fixed prompt, each generating one decision, representing a distinct sampling event}
    \label{1shot}
\end{description}

\textbf{Definition: Few-Shot Querying}:
\begin{description}
    \item{Single API call requesting multiple samples \(n\) with fixed prompt, in the same API call}
    \label{Fewshot}
\end{description}

\begin{enumerate}
    \item \textbf{One-Shot Testing}: Sequential API calls with 1-second delays, alternating between prompts to collect 100 responses per prompt (i.e. 200 API requests) in an independent sampling approach. Each API call is an independent request of the internal model state. 
    
    \item \textbf{Few-Shot Testing}: Single API calls requesting 100 comma-separated responses, with 1-second delays between batches. We collect 10 batches per question, per model. (i.e. 10 API requests only)
    
    \item \textbf{Temperature Impact Analysis}: Temperature settings from 0.5-2.0, collecting 100 responses per setting. We conduct these tests in a one-shot setting, to align with ideal use cases in ABMs for financial decision in line with \cite{vidler2024tradertalkllmbehaviouralabm}.
    
\end{enumerate}
We tested two simple prompts: Q1: "yes or no" and Q2: "Answer randomly, yes or no", designed for broad applicability. Using $\chi^2$ tests, we examined both uniformity and Markovian independence—properties crucial for ABMs. Unintended dependencies in LLM responses could simulate artificial agent interactions, particularly problematic in financial simulations where decision independence is essential.



\subsection{Response Distribution and Independence}
We test models: M1, M2 and M3 with One-shot and Few-Shot sampling methods on Q1 and Q2. We applied chi-square goodness of fit tests (df = 1, \((\alpha = 0.05)\)) to evaluate: \\

    \textbf{Hypothesis 1 (HP1)}: Test for Uniform Distribution of Binary Responses of Yes or No: \\
    \(H_0: p_{\text{yes}} = p_{\text{no}} = 0.5\): responses are uniformly distributed \\
    \(H_1: p_{\text{yes}} \neq p_{\text{no}}\): responses are not uniformly distributed \\
   
    \textbf{Hypothesis 2 (HP2)}: Test for Independence of consecutive decision responses (Markovian responses): \\
    \(H_0: P(Yes_{n}|Yes_{n-1}) = P(Y_{n}) \): the current response, \(n\) is independent of the previous response of \((n - 1)\) \\
    \(H_1: P(Yes_{n}|Yes_{n-1}) \neq P(Yes_{n})\): responses are dependent \\



\section{Results}

\subsection{One-shot testing with fixed Temperature = 1}
\textbf{Unable to reliably replicate a mean of 50\%}: 
In the first test, none of the models accurately replicated the expected distribution (p = 0.5).  While GPT-4o-Mini-2024-07-18's \textbf{Q2} responses still deviated from the expected distribution (43\% yes, 57\% no), it was the only result where we cannot reject the null hypothesis of uniformity and the difference is not statistically significant at the \(\alpha = 0.05\) level (see Table\ref{tab:OneShot_chi_square_tests}). The remaining models (GPT-4-0125-preview and GPT-3.5-Turbo-0125) performed particularly poorly for all questions and were statistically significant deviations away from uniformity (all \(p < 0.001\).  Question 1's response for GPT-4o-Mini-2024-07-18 were also found to be non -uniform.  

\textbf{Model variations are extreme}: across models, with GPT-4-0125-preview producing 99\% "Yes" responses for Q1 compared to GPT-4o-Mini-2024-07-18's 32\%. GPT-4o-Mini showed moderate bias (32-43\% yes), while  GPT-4-0125-preview and GPT-3.5-Turbo-0125 exhibited extreme yes bias (98-99\% and 87-98\% respectively). 

\textbf{Impact of basic prompt with "random" included}: Question 1 simply prompted "yes or no" where as Question 2's "random" framing only improved uniformity in GPT-4o-Mini-2024-07-18.  The \(\chi^2\) statistic reduced from 12.96 to 1.96 (p=0.162), making it the only case where we \textbf{do not reject} the null hypothesis. However, GPT-4-0125-preview and GPT-3.5-Turbo-0125 maintained strong "Yes" biases in Q2 with \(\chi^2 > 87\) and \(p < 1e-20\), regardless of the tested prompts. 

\textbf{Test for Markovian responses}: For \textbf{HP2}, we analyse the Markov property to test response independence, with results reported per model and question in Table \ref{tab:OneShot_markov_tests}. GPT-4-0125-preview and GPT-3.5-Turbo-0125 produce near-perfect \textbf{dependence} in Q1 (P(Yes) = 0.99-1.00, P(Yes|Yes(t-1)) = 0.98-1.00), with deterministic behaviour especially stark in GPT-3.5-Turbo-0125 Q1 responses (86 consecutive Yes→Yes transitions).  While GPT-3.5-turbo-0125 and GPT-4-turbo-preview show low $chi^2$ values (0-0.187), this likely reflects their strong response bias rather than true independence. GPT-4-turbo-0125-preview-mini exhibited more varied responses but showed significant dependencies in both questions ($\chi^2_{Q1} = 21.86$, $\chi^2_{Q2} = 36.19$, both $p < 0.05$), rejecting $H_0$.


\textit{Note: In tables, "accept" \(H_0\) denotes failure to reject the null hypothesis, used for brevity of display.}

\begin{table}[ht]
\small
\begin{tabular}{|l|c|c|c|c|c|c|}
\hline
\textbf{Model} & \textbf{Q} & \textbf{Yes} & \textbf{No} & \textbf{$\chi^2$} & \textbf{p-value} & \textbf{Result} \\
\hline
GPT-\\4o-Mini & 1 & 32 & 68 & 13.0 & 3.18e-4 & Reject $H_0$ \\
& 2 & 43 & 57 & 2.0 & 0.162 & \cellcolor{lightgreen} \textbf{NotReject} $H_0$ \\
\hline
GPT-4 & 1 & 99 & 1 & 96.0 & 1.13e-22 & Reject $H_0$ \\
& 2 & 98 & 2 & 92.2 & 7.99e-22 & Reject $H_0$ \\
\hline
GPT-3.5 & 1 & 87 & 0 & 87.0 & 1.09e-20 & Reject $H_0$ \\
& 2 & 98 & 2 & 92.2 & 7.99e-22 & Reject $H_0$ \\
\hline
\multicolumn{7}{l}{\footnotesize{$H_0$: Responses are uniformly distributed (p = 0.5), $\alpha = 0.05$}} \\
\end{tabular}
\caption{One-Shot Chi-Square Tests for Uniform Distribution}
\label{tab:OneShot_chi_square_tests}
\end{table}

        \begin{table}[ht]
        \small
        \begin{tabular}{|l|c|c|c|c|l|}
        \hline
        \textbf{Model} & \textbf{P(Y)} & \textbf{E[P(Y|Y)]} & \textbf{$\chi^2$} & \textbf{YY/n} & \textbf{Result} \\
        \hline
        GPT-\\ 4o-Mini & 0.32 & 0.28 & 21.9 & 9/32 & Reject $H_0$ \\
        &  0.43 & 0.48 & 36.2 & 20/42 & Reject $H_0$ \\
        \hline
        GPT-4 & 0.99 & 0.99 & 0.04 & 97/98 & Reject$^*$ \\
        & 0.98 & 0.98 & 0.19 & 95/97 & Reject$^*$ \\
        \hline
        GPT-3.5& 1.00 & 1.00 & - & 86/86 & Reject$^*$ \\
        & 0.98 & 0.98 & 0.19 & 95/97 & Reject$^*$ \\
        \hline
        \multicolumn{6}{l}{\footnotesize{$H_0$: P(Yes|Yes$_{t-1}$) = P(Yes), $\alpha = 0.05$}} \\
        \multicolumn{6}{l}{\footnotesize{$^*$Rejected due to near-perfect dependence}} \\
        \end{tabular}
        \caption{One-Shot Markov Property Tests Across Models}
        \label{tab:OneShot_markov_tests}
        \end{table}


\subsection{Few-Shot Results: Improved Performance with Multi-Sample Generation}
Our few-shot testing, where LLMs generated 100 responses per API call across 10 iterations, showed significantly improved uniformity. When averaged, all models and questions approximated the expected 50/50 distribution more closely than in one-shot testing. All cases could \textbf{not} reject the null hypothesis except GPT-4o-Mini-2024-07-18's Q2 responses—notably, the only case that showed uniformity in one-shot testing. Results are detailed in Table \ref{tab:FewShot_Avg_chi_square_tests_part2}.



\begin{table}[ht]
\small
\begin{tabular}{|l|c|c|c|c|c|l|}
\hline
\textbf{Model-Q} & \textbf{Yes} & \textbf{No} & \textbf{$\chi^2$} & \textbf{p-value} & \textbf{Result} \\
\hline
GPT-4o-Mini-Q1 & 527 & 473 & 2.97 & 0.088 & \cellcolor{lightgreen}\textbf{NotReject}$H_0$ \\
GPT-4o-Mini-Q2 & 534 & 466 & 4.62 & 0.032 & Reject $H_0$ \\
\hline
GPT-4-Q1 & 523 & 477 & 2.17 & 0.146 & \cellcolor{lightgreen}\textbf{NotReject}$H_0$ \\
GPT-4-Q2 & 530 & 470 & 3.60 & 0.058 & \cellcolor{lightgreen}\textbf{NotReject}$H_0$ \\
\hline
GPT-3.5-Q1 & 498 & 502 & 0.02 & 0.899 & \cellcolor{lightgreen}\textbf{NotReject}$H_0$ \\
GPT-3.5-Q2 & 498 & 502 & 0.02 & 0.899 & \cellcolor{lightgreen}\textbf{NotReject} $H_0$ \\
\hline
\multicolumn{6}{l}{\footnotesize{$H_0$: Responses are uniformly distributed (p = 0.5), $\alpha = 0.05$}} \\
\end{tabular}
\caption{Few-shot query Chi-Square Tests for Uniform Distribution}
\label{tab:FewShot_Avg_chi_square_tests_part2}
\end{table}


However, the results reveal concerning patterns. Lists of yes/no pairs show strong sequential dependencies, rejecting the Markov property (HP2) across all models. The conditional probability P(Yes|Yes) varies from below 10\% for GPT-3.5-turbo-0125 (Q1) to 28\% and 36\% for GPT-4o-mini-204-07-18 in Q1 and Q2 respectively—lower than one-shot tests. This suggests that when generating sequences, models over-compensate for randomness by excessive alternation, resulting in less random outputs than one-shot responses.

        \begin{table}[ht]
        \small
        \begin{tabular}{|l|c|c|c|c|c|}
        \hline
        \textbf{Model} & \textbf{P(Y)} & \textbf{E[P(Y|Y)]} & \textbf{$\chi^2$} & \textbf{YY/N} & \textbf{\(H_0\)} \\
        \hline
        GPT \\ -4o-Mini& 0.53 & 0.26 & 63.9 & 145/527 & Reject \\
        & 0.53 & 0.36 & 32.5 & 193/533 & Reject \\
        \hline
        GPT-4& 0.52 & 0.27 & 67.2 & 141/523 & Reject \\
        & 0.53 & 0.36 & 31.1 & 191/530 & Reject \\
        \hline
        GPT-3.5& 0.50 & 0.06 & 187.4 & 32/498 & Reject \\
        & 0.50 & 0.35 & 21.7 & 174/497 & Reject \\
        \hline
        \multicolumn{6}{l}{\footnotesize{$H_0$: P(Yes|Yes) = P(Yes), $\alpha = 0.05$, $\chi^2_{crit} = 3.841$}} \\
        \end{tabular}
        \caption{Few-shot query Markov Property Tests Across Models}
        \label{tab:fEWsHOT_Avg_markov_tests_batch}
        \end{table}

\subsection*{Inter-batch Analysis}

\textbf{Mean:} 
Looking at the 10 batches of 100 responses (within 1 API), the average Yes\% is 51.83\% across all runs and the average \(\chi^2\) level is 0.37 with a maximum of any batch across all models being 1.96, well below the test statistic threshold of 3.841, meaning we cannot reject \(H_0\). 

\textbf{Markovian Independence}:
Exploring independence within and across batches, we see that a mixed pattern of results with the main theme being that Question 2 produces more reliable results per batch and again the SOTA GPT-4o-mini-2024-07-18 has the advantage over the other models. Across 10 batches, for 3 models and 2 questions each, we test 60 sets of 100 few-shot responses.  Across these 60 sets we see \(H_0\) is \textbf{not} reject 35 times, and 80\% of Q2 results cannot reject \(H_0\).  Results broken down by model show that GPT-4-0125-preview has the highest number of batches that can not reject \(H_0\) but in terms of robustness to prompts, GPT-4o-Mini-2024-08-17 shows most consistency (across Q1 and Q2), where as other models only Q2 is not rejected.  This mixed pattern of Markovian independence of responses is more promising than the one-shot version where only 1 test was found to be Markovian. However, tests across batches still produced biased distribution in 41.7\% (25 or 60) batch results. 

\subsection*{Limitations of practical implementation of Few-shot methods}
Few-shot sampling in ABMs diverges from realistic individual decision-making by providing multiple decisions simultaneously. This approach increases computational overhead, LLM costs, and memory requirements, while requiring agents to process distribution data statistically. Given these limitations and their impact on model performance, we do not pursue further few-shot analysis in this paper.

The testing of Few-Shot methods across different models (GPT-3.5, GPT-4, and GPT-4o-Mini) failed to reject the null hypothesis in 58.3\% of cases, with a notably higher non-rejection rate for Q2 (80\%) compared to Q1 (36.7\%).


\subsection{Temperature: Impact of Temperature settings on One-shot Query results}
\begin{figure}[htpb]
            \includegraphics[width=0.75\columnwidth]{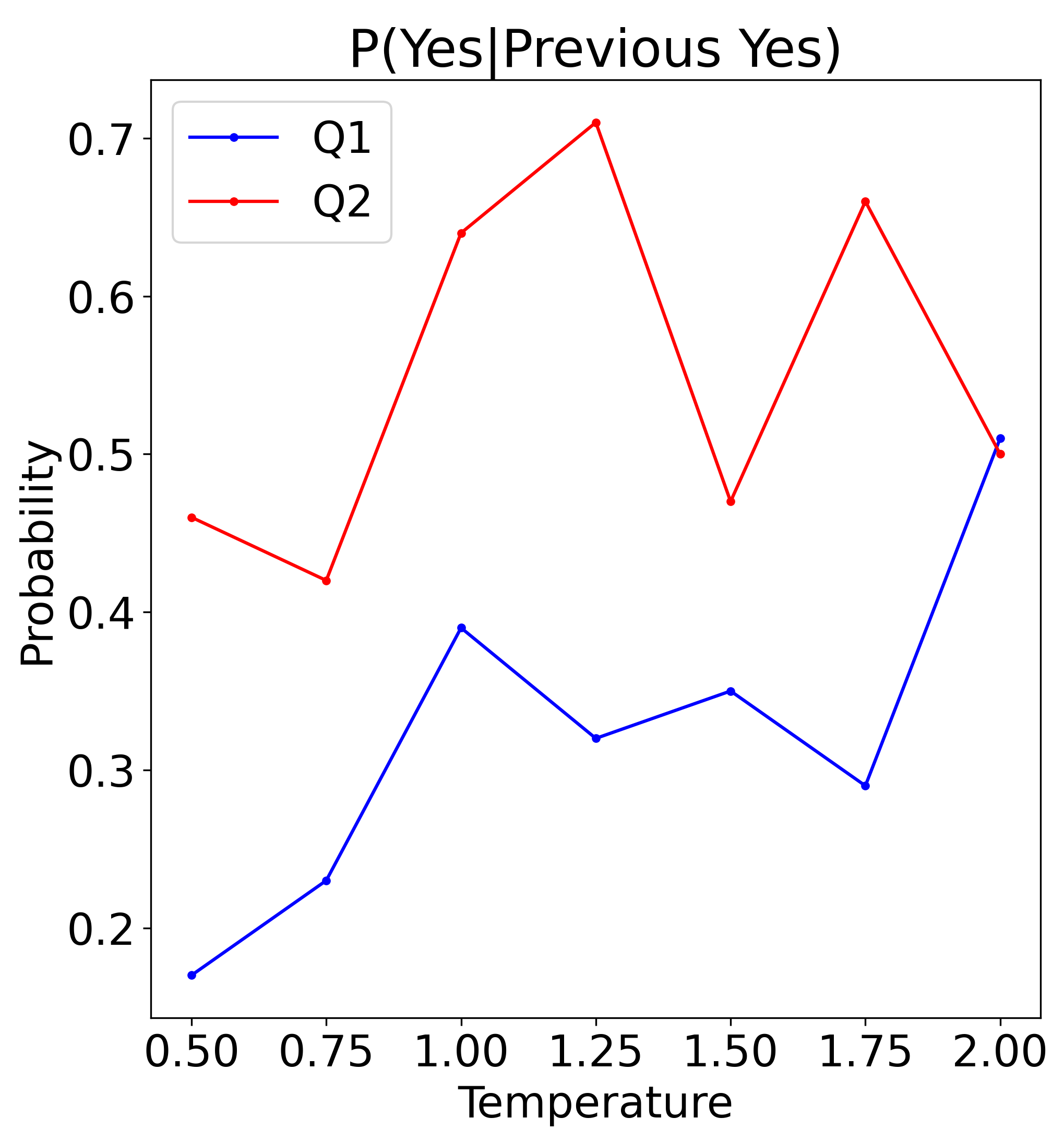}
                    \caption{4o-Mini results for various Temperature settings show non-linear effects}
                    \label{fig:Temp_4oNon linearresults}
        \end{figure}

Exploring temperature settings for One-Shot Query testing at values: \(\mathbb{T}= [0.5, 0.75, 1.0, 1.25, 1.5, 1.75, 2.0] \) we find that LLM APIs do not accept \(T > 2\).  Temperature variations tested in GPT-3.5 and GPT-4-0125-preview consistently yields skewed results, with 73\% to 100\% of responses being "Yes", failing to approximate the expected 50\% mean and showing response dependence, rendering Markovian tests ineffective. No temperature variation tested was able to produce a binary distribution or Markovian response sequence in these models. 
In contrast, varying Temperature in GPT-4o-Mini-2014-07-18 did effects the mean Yes/No levels non-linearly (see Figure \ref{fig:Temp_4oNon linearresults}).  Also, tests at Temperature = 1 reveal variable \(P(Y)\) from previous tests reported earlier in this paper (see Table \ref{tab:temperature_markov_tests}). The only model that produces Markovian response is GPT-4o-Mini-2014-07-18, and it did so across both questions and all temperature settings, in line with our earlier results reported here.

In summary, while temperature adjustments behave as anticipated, they fail to influence the outcomes of our binary distribution in this scenario. Our findings highlight that different model versions can yield significantly varied results.

    \begin{table}[ht]
    \small
    \begin{tabular}{|l|c|c|c|c|}
    \hline
   
    \textbf{Temperature} & \textbf{Q} & \textbf{P(Y)} & \textbf{P(Y|Y)} & \textbf{YY/n} \\
    \hline
    \multicolumn{5}{|l|}{\cellcolor{lightgreen}\textbf{All tests: Failed to reject $H_0$ at $\alpha = 0.05$}} \\
    \hline

    0.50 & Q1 & 0.18 & 0.17 & 3/18 \\
         & Q2 & 0.36 & 0.46 & 16/35 \\
    \hline
    0.75 & Q1 & 0.30 & 0.23 & 7/30 \\
         & Q2 & 0.49 & 0.42 & 20/48 \\
    \hline
    1.00 & Q1 & 0.33 & 0.39 & 13/33 \\
         & Q2 & 0.56 & 0.64 & 35/55 \\
    \hline
    1.25 & Q1 & 0.38 & 0.32 & 12/38 \\
         & Q2 & 0.71 & 0.71 & 50/70 \\
    \hline
    1.50 & Q1 & 0.37 & 0.35 & 13/37 \\
         & Q2 & 0.53 & 0.47 & 25/53 \\
    \hline
    1.75 & Q1 & 0.31 & 0.29 & 9/31 \\
         & Q2 & 0.63 & 0.66 & 40/61 \\
    \hline
    2.00 & Q1 & 0.42 & 0.51 & 20/39 \\
         & Q2 & 0.48 & 0.50 & 22/44 \\
    \hline
    \end{tabular}
    \caption{Temperature Effects on Markov Properties (GPT-4o-Mini)}
    \label{tab:temperature_markov_tests}
    \end{table}

\subsection{Yes/No frequency in Common Crawl open repository of web data}

The best performing model, 4o-Mini-2024-07-18, with a training cut off date of 2024-07-18 prompted an investigation into potential training data biases. Using Common Crawl \cite{commoncrawl} (CC-MAIN-2024-30), we analysed "Yes"/"No" (case insensitive) frequencies in web archives up to the model's training cutoff. We examined both truncated (first 1000 characters) and full page content across samples ranging from 100 to 25,000+ pages, providing insight into potential word frequency biases in LLM training data.  We have no reason to suspect that the results will differ for other web archive scraping methods or sources.

Analysis of Common Crawl data revealed distinct patterns: truncated data showed "Yes" comprising 9.7\% of Yes/No instances (P(Yes|Yes or No) = 33.5\%), with 70\% of pages containing neither term. Full-page analysis found Yes/No responses in 51\% of pages, with P(Yes) = 10\% and P(Yes|Yes or No) = 20.3\%. Chi-square tests (p < 0.05) confirm significant deviation from uniformity across all sample sizes (detailed in Table \ref{tab:commoncrawl_summary}). These findings highlight discrepancies between both the assumed uniform distribution in LLMs (P(Yes) = P(No) = 0.5) and the human usage actual outputs.

\begin{table}[ht]
\small
\begin{tabular}{|l|r|r|r|r|r|}
\hline
\textbf{Pages} & \textbf{Yes} & \textbf{No} & \textbf{Prob.} & \textbf{No} & \textbf{Cond.} \\
& & & & \textbf{Word \%} & \textbf{Prob.} \\
\hline
\multicolumn{6}{|l|}{\textbf{Truncated Data}} \\
\hline
100 & 54 & 606 & 8.2\% & 70\% & 27.3\% \\
1,000 & 57 & 425 & 11.8\% & 75\% & 47.3\% \\
10,000 & 548 & 5,226 & 9.5\% & 69\% & 30.8\% \\
28,632* & 1,676 & 16,131 & 9.4\% & 67\% & 28.6\% \\
\hline
Average & -- & -- & 9.7\% & 70\% & 33.5\% \\
\hline
\multicolumn{6}{|l|}{\textbf{Full Sites}} \\
\hline
100 & 70 & 1,287 & 5.2\% & 57\% & 12.0\% \\
1,000 & 789 & 6,628 & 10.6\% & 52\% & 22.3\% \\
10,000 & 16,215 & 110,621 & 12.8\% & 49\% & 25.1\% \\
26,705* & 47,107 & 361,777 & 11.5\% & 48\% & 22.0\% \\
\hline
Average & -- & -- & 10.0\% & 51\% & 20.3\% \\
\hline
\end{tabular}
\caption{Common Crawl Analysis of Yes/No Occurrences with truncated date to 1000 characters and also full sites (* denotes Rate limited response)}
\label{tab:commoncrawl_summary}
\end{table}


\subsection{Comparing LLM Randomness to Human and True Random Sequences}
Extending recent work by \cite{Harrison} and \cite{Angelike2024}, we compare our one-shot and few-shot results against true random binary sequences from Random.org\footnote{https://www.random.org/integers/?mode=advanced}. Following \cite{BaenaMirabete2023AMT}, we analyze sequences using sliding windows of size $w$ \( (1 \leq w \leq 5)\), where $w$ represents the sequence length influencing subsequent choices. For a binary sequence $S = (s_1, ..., s_n)$ where $s_i \in {\text{Yes}, \text{No}}$, we calculate negative recency effects. Our method is as follows:

\subsubsection{Baseline Switching Rate}
The baseline switching rate represents the overall probability of alternation in the sequence.  

    \begin{equation}
    \text{baseline\_rate} = \frac{\sum_{i=1}^{|S|-1} \mathbb{1}_{s_i \neq s_{i-1}}}{|S|-1}
    \end{equation}
    
    where $\mathbb{1}_{s_i \neq s_{i-1}}$ is the indicator function that equals 1 if adjacent elements differ and 0 otherwise.
    
\subsubsection{Window-Specific Switching Rates}
For each window size $w$, we calculate the switching rate after runs of length $w$ as:

    \begin{equation}
    \text{switch\_rate}_w = \frac{\sum_{i=w}^{|S|-1} \mathbb{1}_{s_i \neq s_{i-1}} \cdot \mathbb{1}_{\text{run}_w(i)}}{|\{i : \text{run}_w(i)\}|}
    \end{equation}

where:
    \begin{itemize}
        \item $\mathbb{1}_{\text{run}_w(i)}$ equals 1 if position $i$ follows a run of length $w$
        \item $|\{i : \text{run}_w(i)\}|$ is the number of positions that follow runs of length $w$
    \end{itemize}

\subsubsection{Recency Effect} 
The recency effect for window size $w$ is then calculated as the difference between the window-specific switching rate and the baseline rate; 
    \begin{equation}
    \text{recency\_effect}_w = \text{switch\_rate}_w - \text{baseline\_rate}
    \end{equation}
    
A positive recency effect indicates negative recency (increased tendency to switch after runs), while a negative value indicates positive recency (decreased tendency to switch after runs).

\textbf{Hypothesis 3 (HP3)}: We test a Two-Sample t-Test, comparing switching rates between two sequences of binary outcomes for Recency bias. The first sequence is an LLM output, the second is a binary series created by Random.org's binary generator. To reject $H_0$ is to state that the true random binary sequence has a different patten to the LLM output and thus, the LLM output contains recency bias: \\

\(H_0: \mu_1 = \mu_2\) : The mean switching rates after runs are equal between sequences, LLM is indistinguishable from a true random binary sequence \\
\(H_1: \mu_1 \neq \mu_2\) : The mean switching rates after runs differ between sequences





\subsection{Recency Bias Results}
Analysis reveals that most model outputs exhibit recency bias compared to true random sequences, with few exceptions (highlighted in green). While GPT-4o-Mini-2024-07-18 avoided the human recency effects documented by \cite{BaenaMirabete2023AMT} for 1 and 3-step lookbacks.  In this way, these model combinations could be considered more "random" than humans, however these models still failed to simultaneously achieve both uniform distribution and Markovian independence.

    \begin{table}[ht]
    \small
    \begin{tabular}{|l|r|r|r|r|r|r|}
    \hline
    \textbf{Model} & \multicolumn{3}{c|}{\textbf{1-shot}} & \multicolumn{3}{c|}{\textbf{FewShot}} \\
    \cline{2-7}
    & \textbf{RNG} & \textbf{M} & \textbf{Mk} & \textbf{RNG} & \textbf{M} & \textbf{Mk} \\
    \hline
    4o-Mini Q1 & \cellcolor{lightgreen} A,  \(P>1\) & R & R & R=NR & \cellcolor{lightgreen}A & R \\
    \hline
    4o-Mini Q2 & \cellcolor{lightgreen} A & \cellcolor{lightgreen}A & R & R=NR & R & R \\
    \hline
    GPT-4 Q1 & R=NR & R & R & \cellcolor{lightgreen}A P=4 & \cellcolor{lightgreen}A & R \\
    \hline
    GPT-4 Q2 & R=NR & R & R & \cellcolor{lightgreen}A P=4 & \cellcolor{lightgreen}A & R \\
    \hline
    GPT-3.5 Q1 & R=NR & R & R & R=NR & \cellcolor{lightgreen}A & R \\
    \hline
    GPT-3.5 Q2 & R=NR & R & R & \cellcolor{lightgreen}A P=4,5 & \cellcolor{lightgreen}A & R \\
    \hline
    \end{tabular}
    \caption{Statistical Analysis of Randomness Tests (R=reject, A=Not Reject\(H_0\), NR=Non random, M=Mean, Mk=Markov). Green cells indicate Non-Rejection \(H_0\).}
    \label{tab:model_randomness}
    \end{table}


\section{Conclusion}
We test three LLM model subversions for decision making biases by examining binary decision outputs across: GPT-3.5 Turbo-0125, GPT-4-0125-preview, and GPT-4o-Mini-2034,07,18.   We find they cannot adequately replicate a uniform distribution in independent sampling of these models (One-Shot).   

We also find statistically significant performance variations between models (GPT 4 to GPT 3.5) and especially between specific sub-versions (GPT 4 and 4o-Mini), in addition to significant impacts of sampling methods on results. Using a simple Yes/No benchmark task, only GPT-4o-Mini-2024-07-18 (one-shot) achieves output not statistically different from uniform distribution, though responses remain non-Markovian. Temperature adjustments (0.5-2.0) failed to reliably influence these distributions, with GPT-4o-Mini showing non-linear responses while other models maintained strong biases regardless of temperature setting.
 
The few-shot methodology, while producing better distributional outcomes, are less practical for ABM applications and nearly half of all tests result in non-Markovian decision sequences, indicating persistent temporal dependencies. Between model versions, we observe substantial variations in response patterns, suggesting architectural and training differences significantly impact decision making capabilities. We further compare results to true random binary series and test specifically for the common human bias of Negative Recency - finding LLMs have a mixed ability to 'beat' humans, with GPT-4o-Mini-2024-07-18 notably avoiding human recency effects in one-shot testing, though still producing non-Markovian outputs.

These findings expose systematic biases in LLM-based decision making, sensitivity to model sub-versions and to sampling methods. These hold critical implications for ABMs, particularly in finance where Markovian properties are typically assumed. 

Building on these findings, future work should focus on several key areas: enhancing LLM integration in financial ABMs through improved real-time decision mechanisms; investigating the relationship between model architecture, underlying causes of bias and decision-making capabilities; evaluating performance across non-OpenAI models; exploring methods to mitigate identified decision biases; and examining implications for broader financial modelling applications. This research pathway aims to better understand and address the challenges of implementing LLMs in practical financial modelling environments, ultimately improving their reliability for real-world applications.



\section{Acknowledgements}
This work is funded in part by an ARC Laureate grant FL200100204 and NSF-CSIRO grant to Toby Walsh.

\bibliography{bib}

\end{document}